\documentclass{article} 
\usepackage{iclr2017_conferenceA,times}
\usepackage{hyperref}
\usepackage{url}

\title{Representation Stability as a Regularizer for Improved Text Analytics Transfer Learning
}

\author{Matthew Riemer, Elham Khabiri, and Richard Goodwin \\
IBM T.J. Watson Research Center\\
Yorktown Heights, NY, USA \\
{\texttt \{mdriemer, ekhabiri, rgoodwin\}@us.ibm.com}}

\begin{document}

\maketitle

\begin{abstract}

Although neural networks are well suited for sequential transfer
learning tasks, the catastrophic forgetting problem hinders proper
integration of prior knowledge. In this work, we propose a solution to
this problem by using a multi-task objective based on the idea of
distillation and a mechanism that directly penalizes forgetting at the
shared representation layer during the knowledge integration phase of
training. We demonstrate our approach on a Twitter domain sentiment analysis
task with sequential knowledge transfer from four related tasks.  We
show that our technique outperforms networks fine-tuned to the target
task. Additionally, we show both through empirical evidence and examples that it does not forget useful knowledge from the source task that is forgotten during standard fine-tuning.
Surprisingly, we find that first distilling a human made rule based
sentiment engine into a recurrent neural network and then integrating
the knowledge with the target task data leads to a substantial gain in
generalization performance. Our experiments demonstrate the
power of multi-source transfer techniques in practical text analytics
problems when paired with distillation. In particular, for the SemEval
2016 Task 4 Subtask A \citep{SemEval} dataset we surpass the state of the
art established during the competition with a comparatively simple
model architecture that is not even competitive when trained on only
the labeled task specific data.

\end{abstract}

\section{Introduction}

Sequential transfer learning methodologies leverage knowledge representations from a source task in order to improve performance for a target task. A significant challenge faced when transferring neural network representations across tasks is that of catastrophic forgetting (or catastrophic interference). This is where a neural network experiences the elimination of important old information when learning new information. The very popular strategy of fine-tuning a neural network involves first training a neural network on a source task and then using the model to simply initialize the weights of a target task network up to the highest allowable common representation layer. However it is highly susceptible to catastrophic forgetting, because in training for the target task it has no explicit incentive to retain what it learned from the source task. While one can argue that forgetting the source task should not matter if only the target task is of interest, our paper adds to the recent empirical evidence across problem domains \citep{LwF},\citep{PNN} that show additional network stability can lead to empirical benefits over the fine-tuning algorithm. It seems as though for many Deep Learning problems we can benefit from an algorithm that promotes more stability to tackle the well known stability-plasticity dilemma. One popular approach for addressing this problem is rehearsals \citep{Murre92}, \citep{Robins95}. Rehearsals refers to a neural network training strategy where old examples are relearned as new examples are learned. In the transfer setting it can be seen as related to multi-task learning \citep{Caruana97} where two tasks are trained at the same time, rather than sequentially, while sharing a common input encoder to a shared hidden representation. However, in rehearsals the representation is biased in favor of the source task representation through initialization. This technique is very sensible because while fine-tuning is susceptible to catastrophic forgetting, multi-task learning is not \citep{Caruana97}. 

One of the biggest issues with the standard rehearsals paradigm is that it requires a cached memory of training examples that have been seen in the past. This can be a massive requirement as the number of source tasks and training data sizes scale. One compelling technique for addressing this problem is the concept of pseudorehearsals \citep{Robins95}, \citep{Robins96}, where relearning is performed on an artificially constructed population of pseudoitems instead of the actual old examples. Unfortunately, current automatic techniques in the text analytics domain have not yet mastered producing linguistically plausible data. As such, the pseudorehearsals paradigm is likely to waste computational time that could be spent on learning realistic patterns that may occur during testing. In our work, we extend the Learning without Forgetting (LwF) paradigm of \citep{LwF} to the text analytics domain using Recurrent Neural Networks. In this approach, the target task data is used both for learning the target task and for rehearsing information learned from the source task by leveraging synthetic examples generated for the target task input by the model that only experienced training on the source task data. As argued by \cite{LwF}, this setup strikes an important balance between classification performance, computational efficiency, and simplicity in deployment.   

Regardless of whether they are applied to real source task examples, real target task examples, or synthetic examples, paradigms in the style of rehearsals all address the shortcomings of neural network forgetting by casting target task integration as a multi-task learning problem. However, this is not quite the purpose of the multi-task learning architecture, which was designed for joint learning of tasks from scratch at the same time. The key disconnect is that in multi-task learning, the transformation from the shared hidden layer to the outputs for each task are all learned and updated with the changing hidden representation. This would imply that, in the framework of rehearsals, it is possible for there to be significant changes during learning of the network's representation, and thus its abilities on the source task itself. While it would be desirable to claim we were allowing our source task network to become even better based on the target task than it was before, this motivation seems idealistic in practice. One reason this is idealistic is because multi-task learning generally only works well when tasks are sampled at different rates or alternatively given different priority in the neural network loss function \citep{Caruana97}. As a result, it is most likely that auxilirary source tasks will receive less priority from the network for optimization than the target task. Additionally, we observe in our experiments, and it has been observed by others in \citep{MTD}, that it is generally not possible to distill multiple complex tasks into a student network at full teacher performance for all tasks. This seems to imply the degradation of the source task performance during training is somewhat inevitable in a multi-task learning paradigm.

We address this issue with our proposed forgetting cost technique. We demonstrate that it, in fact, can be valuable to keep the hidden to output transformation of the source tasks fixed during knowledge integration with the target task. This way, we impose a stronger regularization on the hidden representation during target task integration by not allowing it to change aspects that were important to the source task's performance without direct penalization in the neural network's loss function. We demonstrate empirically both that freezing the source task specific weights leads to less deterioration in the accuracy on the source task after integration, and that it achieves better generalization performance in our setting. The forgetting cost is practical and easy to implement in training any kind of neural network. In our experiments, we explore application of the forgetting cost in a recurrent neural network to the three way Twitter sentiment analysis task of SemEval 2016 Task 4 Subtask A and find it to achieve consistently superior performance to reasonable baseline transfer learning approaches in four examples of knowledge transfer for this task.

We also demonstrate how powerful distillation can be in the domain of text analytics when paired with the idea of the forgetting cost. Significantly, we show that a high quality gazetteer based logical rule engine can be distilled using unlabeled data into a neural network and used to significantly improve performance of the neural network on the target task. This is achieved with a novel extension of the LwF paradigm by \cite{LwF} to the scenario of a source task with the same output space as the target task. This can be a very promising direction for improving the ability of humans to directly convey knowledge to deep learning algorithms. Indeed, a human defined rule can contain far more information than a single training example, as that rule can be projected on to many unlabeled examples that the neural network can learn from. This is the reason human teachers generally begin teaching human students tasks by going over core rules at the onset of learning.  Moreover, we showcase that multiple expert networks trained on the target task with prior knowledge from different source tasks can be effectively combined in an ensemble and then distilled into a single GRU model \citep{Cho14}, \citep{Chung14}. Leveraging this combination of distillation and knowledge transfer techniques allows us to achieve state of the art accuracy on the SemEval task with a model that performs 11\% worse than the best prior techniques when trained only on the labeled data.  

\section{Related Work} 

Since the work of \citep{Caruana06} and \citep{Hinton15} showed that an ensemble of neural network classifier can be distilled into a single model, knowledge distillation from a teacher network to a student network has become a growing topic of neural network research. In \citep{Ba14} it was shown that a deep teacher neural network can be learned by a shallow student network. This idea was extended in \citep{FitNets}, where it was demonstrated that a deep and narrow neural network can learn a representation that surpasses its teacher. The use of distillation as a means of sharing biases from multiple tasks was explored in \citep{Vapnik16}, where the teacher network is trained with the output of the other tasks as input. It is not obvious how to extend a recurrent neural network to best use this kind of capability over a sequence. The idea of distilling from multiple source task teachers into a student network was highlighted in the reinforcement learning setting in \citep{MTD}. Additionally, the concept of using distillation for knowledge transfer was also explored in \citep{Net2Net}, where function preserving transformations from smaller to bigger neural network architectures were outlined. This technique could also provide value in some instances for our approach where wider or deeper neural networks are needed for the task being transferred to than was needed for the original task. Distillation over target task data was first proposed as a means of elevating catastrophic forgetting in sequential knowledge transfer as applied to image classification in \citep{LwF}. We extend this approach for its first application to our knowledge for text analytics problems, with a recurrent neural network architecture, and in the setting where the source task and target task have the same output. The chief distinction of our proposed forgetting cost is that source task specific parameters are held fixed during integration with the target task as opposed to the joint training of all parameters used by \cite{LwF}. Our experiments empirically support the intuition that freezing these parameters leads to greater retention of source task performance after target task integration and better generalization to the target task.  

An ensemble over multiple diverse models trained for the same sentiment analysis task was also considered in \citep{ICLR15} for the IMDB binary movie reviews sentiment dataset \citep{IMDB}. We tried this ensemble model in our work and found that it gave very limited improvement. Our ensemble technique learns a more powerful weighted average based on the soft targets of each task and a multi-step greedy binary fusion approach that works better for the Twitter sentiment analysis task in our experiments. Knowledge transfer from multiple tasks was considered to estimate the age of Twitter users based on the content of their tweets in \citep{Riemer15}. We experimented with the hidden layer sharing approach outlined in that work and found that even when using just a single softmax combining layer, it would overfit on our limited training and validation data. Progressive neural networks \citep{PNN} is a recently proposed method very similar in motivation to our forgetting cost as it is directly trying to solve the catastrophic forgetting problem. The idea is that learned weight matrices relate the fixed representations learned on the source task to the construction of representations for the target task. In our experiments, the progressive neural networks approach consistently fails to even match the results achieved with fine-tuning. We hypothesize that although using fixed representations to aid learning addresses catastrophic forgetting, it suffers from the curse of dimensionality. As such, when training data is relatively small given the complexity of the task, it is prone to overfitting as it effectively increases the input dimension size through shared fixed representations. 

The combination of logic rules and neural networks has been explored in a variety of different architectures and settings. These neural-symbolic systems \citep{Garcez2012} include early examples such as KBANN \citep{Towell} that construct network architectures from given rules to perform reasoning. \citep{Hu2016} very recently also looked at the problem of distilling logical rules into a neural network text analytics classifier. However, our approach is much more generic as it can be applied to integrate knowledge from any kind of pre-made classifier and treats the rule engine as a black box. In \citep{Hu2016} they consider the individual rules and leverage an iterative convex optimization algorithm alongside the neural network to regularize the subspace of the network. In our work we demonstrate that, by guarding against catastrophic forgetting, it is possible to efficiently leverage rules for transfer by utilizing a generic sequential knowledge transfer framework.  We do not need to make any modification to the architecture of the neural network during testing and do not need iterative convex optimization during training.

\section{Forgetting Cost Regularization} \label{fc}
\subsection{Sequential Knowledge Transfer Problem Statement} \label{seq}

In the sequential knowledge transfer problem setting explored in this paper, training is first conducted solely on the source task examples $S$, including $K_S$ training examples $(x_{Si},y_{Si}) \in S$ where $x_{Si}$ is the input representation and $y_{Si}$ is the output representation. After training is complete on $S$, we would like to now use prior knowledge obtained in the model trained on $S$ to improve generalization on a new target task with examples $T$, which includes $K_T$ training examples $(x_{Ti},y_{Ti}) \in T$. Here we assume that the input representations $x_{Si}$ and $x_{Ti}$ are semantically aligned in the same representation space. As such, if there is useful knowledge in $S$ that applies in some direct or indirect way to the target task that is not present in $T$, we would expect a good knowledge integration approach to generalize better to the target task than it is possible to using the training data in $T$ alone. Strong performance for the sequential knowledge transfer problem is a first step towards the greater goal of a mechanism for effective lifelong learning \citep{LML}.

\subsection{Forgetting Cost for Tuning a Target Task Model} \label{ft}

The most straightforward application of our proposed forgetting cost paradigm is for the case of integrating a neural network that has been trained on source task data $S$, which has outputs in the same representation space as the outputs for the target task data $T$. In this case, the forgetting cost amounts to the addition of a regularization term in the objective function during the integration phase when we train using $T$. This promotes the neural network to be able to recreate the soft labels of the initialized model found after training on $S$ before integration is started with $T$. More formally:

\begin{equation}
Loss = L(y,\hat y) + \alpha_f L(y_{init},\hat y)
\label{eq:L1}
\end{equation}
where $L$ is some loss function (we use mean squared error in our experiments) and $y_{init}$ is the soft label generated for the target task input $x_{Ti}$ based on the model after training just on $S$. The model trained just on $S$ is also used to initialize the weights of the target task model before integration with $T$ as we do in the standard fine-tuning paradigm.  $\alpha_f$ is a hyperparameter that can be utilized to control the extent of allowed forgetting. Of course, a very similar way to express this idea would be to mix synthetic training examples $T'$ with the same input as $T$ and output generated by the model trained just on $S$ with the true target task training examples $T$. In this case, the mixing rate of the teacher generated training examples is analogous to our forgetting parameter $\alpha_f$ determining the prioritization. These techniques perform quite similarly in our experiments, but we actually find that the formulation in equations \ref{eq:L1} and \ref{eq:L11} perform slightly better on the test set. For example, this formulation is superior by 0.4\% accuracy in tuning a distilled representation of a logical rule engine. We conjecture that learning tasks in the same gradient step when they are related to the same input data results in slightly less noisy gradients.  

\subsection{Forgetting Cost for Knowledge Transfer from a Related Task}\label{fcs}

The assumption in section \ref{ft} that the output of the source task data $S$ should be in the same representation space as the output for the target task data $T$ is quite a big one. It rules out the vast majority of knowledge sources that we can potentially leverage. As such, we propose an extension that does not make this restriction for application in sequential knowledge transfer of tasks that are not directly semantically aligned. We update our model to include another predicted output separate from $\hat y$:

\begin{equation}
\hat y_{init} = f_{init}(W_{fixed}h_{shared} + b_{fixed})
\label{eq:yinit}
\end{equation}

where $\hat y_{init}$ is a predicted output attempting to recreate the soft labels of the original model trained just on $S$. $f_{init}$ is the non-linearity used in the final layer of the source task model. Weight matrix $W_{fixed}$ and bias $b_{fixed}$ are taken from the final layer of the source task model and are not updated during integration with the target task data $T$. As a result, the loss function is updated from section \ref{ft}:

\begin{equation}
Loss = L(y,\hat y) + \alpha_f L(y_{init},\hat y_{init})
\label{eq:L11}
\end{equation}
where the hidden state is shared between both terms in the objective function. Up to the shared hidden layer, we initialize the model for the target task with the weights learned just using $S$. Random matrices and bias vectors are now used to initialize the prediction of $\hat{y}$ based on the shared hidden representation. This can be seen as a weak form of restricting the model parameters that can be useful for regularization. The hidden representation is in effect constrained so that it is promoted not to change in key areas that have a large effect on the output vector of the source task model. On the other hand, there is little regularization for parameters that have little effect on the output vector for the source task model.

\section{Recurrent Neural Network Model} \label{GRU}

In recent years, recurrent neural network models have become a tool of choice for many NLP tasks. In particular, the LSTM variant \citep{LSTM} has become popular as it alleviates the vanishing gradients problem \citep{Bengio94} known to stop recurrent neural networks from learning long term dependencies over the input sequence. In our experiments we use the simpler GRU network \citep{Cho14}, \citep{Chung14} that generally achieves the same accuracy despite a less complex architecture. Each time step $t$ is associated with an input $x_t$ and a hidden state $h_t$. The mechanics of the GRU are defined with the following equations:

\begin{equation}
z_t = \sigma(W_{xz} x_t + W_{hz} h_{t-1})
\label{eq:z_t}
\end{equation}
\begin{equation}
r_t = \sigma(W_{xr} x_t + W_{hr} h_{t-1})
\label{eq:r_t}
\end{equation}
\begin{equation}
\tilde h_t = tanh(W_{xh} x_t + r_t \circ W_{hh} h_{t-1})
\label{eq:h1_t}
\end{equation}
\begin{equation}
h_t = z_t \circ h_{t-1} + (1-z_t) \circ \tilde h_t
\label{eq:h2_t}
\end{equation}

where $\circ$ denotes an element-wise product. $W_{xz}$, $W_{xr}$, and $W_{xh}$ represent learned matrices that project from the input size to the hidden size. $W_{hz}$, $W_{hr}$, and $W_{hh}$ represent learned matrices that project from the hidden size to the hidden size. In our work we evaluate the GRU in the categorical prediction setting. For each document, the hidden state after the last word $h_L$ is used for the prediction $ \hat y$ of the label $y$. As such, we treat $h_L$ as the shared hidden representation $h_{shared}$ from section \ref{fcs} for our experiments. 

\begin{equation}
\hat y = f(W_{yh}h_L + b_y)
\label{eq:y}
\end{equation}

The prediction goes through one other non-linear function $f$ after the final hidden state is derived. In our experiments we use the softmax function, but others are useful in different settings. A model that builds on top of GRUs with an external memory storage paradigm \citep{Socher15} currently holds the state of the art on movie review sentiment analysis. However, we focus just on the straightforward single layer GRU model in our experiments so that we can more easily disentangle factors of influence on performance. Our GRU model was fed a sequence of fixed 300 dimensional Glove vectors \citep{Glove}, representing words based on analysis of 840 billion words from a common crawl of the internet, as the input $x_t$ for all tasks. It has been shown in a number of papers that tuning the word embeddings during training could increase performance, and it is possible our approach could have performed better had we done so.

\section{Sequential Knowledge Transfer Experiments}

\subsection{Experiment Details}

Our neural network models were implemented in Theano \citep{Theano} and trained with Stochastic Gradient Descent. As we did not use an advanced optimization method and noticed run to run variation in performance, for all of our transfer learning models we trained 10 parallel versions and chose the one with the highest validation accuracy. The SemEval 2016 Task 4 Subtask A training set consists of 10,000 total training examples, but we were only able to receive 8,906 because of tweet removals when we used the downloading script. For the target task data across our experiments, 7,600 examples of the SemEval training set examples were used for training and the rest for validation. The GRU model achieves only 53.6\% accuracy on the SemEval testing data when just training with the target task data and random initialization. In order to improve, we consider knowledge transfer from GRUs trained for the following source tasks to the SemEval target task data:

\textbf{Distilling Logical Rules:} Knowledge distillation can be performed using teacher models that are very different in structure than their neural network based student models. We demonstrate with this task that a compilation of logical linguistic rules can be used as an effective teacher for a GRU by having the GRU attempt to create the output of the rule engine generated over unlabeled in domain data. Specifically, our gazetteer based logical rule engine separates sentences and phrases in the text. It then applies dictionaries of positive and negative sentiment words and phrases to the corresponding text.  For each positive or negative phrase found, it checks to see if negation or double negation are applied, and modifies the polarity of the sentiment accordingly. The result for any piece of text is a count of positive and negative sentiment occurrences. For this task, we simply count the total number of positive and negative indicators to give an overall positive, negative or neutral score. We provide addition details on how we mapped rules to soft targets for the student network to recreate in Appendix \ref{AppendixA}. We utilized a GRU model with 50 hidden units and 50,000 unlabeled examples for our source task model. We distill off the soft labels as in \citep{Hinton15}, but set our temperature fixed at 1.0. It is possible that our performance could have improved by tuning this parameter. Additional details about the selection of the network and data size are included in Appendix \ref{AppendixB}. The logical rule model itself achieves 57.8\% accuracy on the SemEval testing data and the rules distilled into a GRU as explained in section \ref{GRU} achieves 58.9\% accuracy before any integration with the SemEval target task data. We leverage this task for comparison of knowledge transfer techniques when the source task and target task share an output space as discussed in section \ref{ft}.

\textbf{Binary Movie Reviews:} For knowledge transfer from related tasks as discussed in section \ref{fcs} we first consider the Stanford Sentiment Treebank \citep{Socher13}, which is a popular sentiment dataset based on the movie review domain. We consider one source task to be the binary (positive, and negative) sentence level sentiment subtask which contains 6,920 training examples, 872 validation examples, and 1,821 testing examples. Our GRU model with 40 hidden units achieves 85.5\% accuracy on this task. 

\textbf{Five Class Movie Reviews:} We also consider another source task leveraging the Stanford Sentiment Treebank data from the fine grained (very positive, positive, neutral, negative, and very negative) sentence level sentiment substask which contains 8,544 training examples, 1,101 validation examples, and 2,210 testing examples. We use a GRU model with 200 hidden units to accommodate for the increased task complexity and achieve 45.9\% accuracy. This fine grained model can actually be assessed directly on the SemEval task by projecting from five classes to three classes, but it only achieves 44.2\% accuracy with no tuning on the target task data. Our performance on these two movie review source tasks is quite similar to what was reported in \citep{TreeLSTM} when using a similar setup, but with LSTMs for both subtasks. 

\textbf{Emoticon Heuristic:} Finally, we consider a semi-supervised task based on emoticon prediction motivated by the successful work in \citep{Go2009}, leveraging it in the twitter sentiment domain and its use as a vital component of the SemEval competition winning system \citep{SwissCheese}. We find unlabelled tweets that contain smileys, frowns, or laughing emoticons. We remove emoticons from the tweet before prediction and compile a dataset of 250,000 training examples, 50,000 validation examples, and 100,000 testing examples for each of the three classes. This is multiple orders of magnitude smaller than the 90 million tweets used in \citep{SwissCheese} to allow for quick experimentation. Our GRU model with 50 hidden units achieves 63.4\% accuracy on the emoticon prediction test set.

\subsection{Sequential Knowledge Transfer Algorithms}

We consider multiple sequential knowledge transfer algorithms for experimental comparison. Each uses only the source task data for learning the source task and only the target task data for integrating with the target task. This way integration is fast and simple, because it does not incorporate storage and replay of examples from the potentially very large source task as argued in \citep{LwF}.

\textbf{Fine-Tuning:} The representation is simply initialized with the representation found after training on the source task and then trained as usual on the target task. This approach was pioneered in \citep{Hinton2006}, in application to unsupervised source tasks and applied to transfer learning in \citep{Bengio2012}, and \citep{Mesnil2012}. The learning rate is tuned by a grid search based on the validation set performance.

\textbf{Progressive Networks:} We also compare with our implementation of a progressive neural network \citep{PNN}, where the representation learned for the source task is held fixed and integrated with a target task specific model via lateral connections trained using the target task data. The learning rate is also tuned based on a grid search using the validation set.

\textbf{Learning without Forgetting (LwF):} In the LwF paradigm, joint training is performed after parameter initialization. This is achieved by treating the target task data and the output generated by the source task model based on the target task input data as two jointly learned tasks as in \citep{Caruana97}. As opposed to our proposed forgetting cost, the source task specific parameters are not held fixed while training on the target task data. The learning rate and mixing rate between the tasks are tuned by a grid search based on validation set performance. We first consider a version of the LwF model that leverages a random initialization of the target task specific parameters and initialization of all parameters learned on the source task with the learned values. We also consider another formulation that we call Greedy LwF. This is actually more closely aligned with the original paper \citep{LwF}. All source task parameters are first held fixed, and the target task specific parameters are learned alone before joint training with all of the parameters unfrozen as a second step. For the case of source tasks with output in the space of the target task output, there are no source task specific parameters, so the forgetting cost can be viewed as a viable interpretation of the LwF paradigm appropriate in that setting. 

\textbf{Forgetting Cost:} Finally, we compare each baseline model with our proposed forgetting cost described in section \ref{fc}. The learning rate as well as $\alpha_f$ from equations \ref{eq:L1} and \ref{eq:L11} were tuned by a grid search based on the validation set performance.

\subsection{Target Task Results}

We empirically evaluate the generalization performance of the forgetting cost for sequential knowledge transfer from four different source tasks in Table \ref{tab111} and Table \ref{tab1111}. The source task considered in Table \ref{tab111} is distilling a logical rule model, leveraging the technique outlined in equation \ref{eq:L1}. In Table \ref{tab1111} we leverage the forgetting cost for related task knowledge transfer as outlined in equation \ref{eq:L11}.  

Our experimental results on the SemEval data validate our intuition that the forgetting cost should lead to stronger regularization and better generalization performance. One thing to note about our progressive neural networks implementation is that it effectively has only one hidden layer, because we hold our embeddings fixed during model training and the same embeddings are shared among the models used for all of the tasks. It is possible that having multiple layers of lateral connections is important to achieving good performance. However, this setting was not applicable in our experiments. Our results for sequential knowledge transfer on the SemEval benchmark are quite encouraging as the forgetting cost outperforms baselines significantly in all cases. 

We additionally have validated the intuition that equation \ref{eq:L1} should perform stronger regularization than equation \ref{eq:L11} when equation \ref{eq:L1} is applicable. In fact, for our distilled logical rule model tuning experiments, we found that equation \ref{eq:L1} performs 3\% better on the test set. In an attempt to understand more about what caused this performance difference, we monitored testing set performance at each epoch and noticed that equation \ref{eq:L11} is actually prone to overfitting away from a good solution on the test set. However, it often finds a pretty good one comparable to equation \ref{eq:L1} early in training. When equation \ref{eq:L1} could be applied, it seems to be a useful regularization to constrain both the hidden layer and the output layer to align with the model learned on the source task. In equation \ref{eq:L11}, the hidden to output transformation learned for the target task can in contrast learn to deviate from the transformation learned for the source task.  

\begin{table}
\small
\centering
\begin{tabular}{|l|c|}
\hline \bf Model Description & \bf Accuracy on SemEval Test Set \\ \hline
Forgetting Cost Transfer & \textbf{64.4\%} \\
Fine-tuning Transfer & 58.5\% \\
Progressive Networks Transfer & 56.9\% \\ \hline
Distilled Logical Rule Model & 58.9\% \\
Logical Rule Model & 57.8\% \\ 
GRU Trained on Only SemEval Data & 53.6\% \\ \hline
\end{tabular}
\caption{\label{tab111} Evaluation of target task tuning methodologies for a distilled rule model to the task of SemEval 2016 Task 4 Subtask A.}
\end{table}

\begin{table}
\small
\centering
\begin{tabular}{|l|c|c|c|c|c|}
\hline \bf Source Task & \bf Fine-Tuning & \bf Progressive Networks & \bf LwF & \bf Greedy LwF & \bf Forgetting Cost \\ \hline
Binary Movie Reviews & 57.3\% & 54.5\% & 58.1\% & 58.8\% &  \textbf{59.7\%} \\ \hline
Five Class Movie Reviews & 57.4\% & 54.6\% & 57.1\% & 56.6\% &  \textbf{58.2\%} \\ \hline
Emoticon Heuristic & 55.8\% & 53.2\% & 57.7\% & 56.7\% & \textbf{58.6\%}\\ \hline
\end{tabular}
\caption{\label{tab1111} Evaluation of knowledge transfer from three source tasks to the task of SemEval 2016 Task 4 Subtask A.}
\end{table}

\subsection{Source Task Performance After Target Task Integration}

In Table \ref{tabv} we explore the retention of empirical performance on the source task for knowledge transfer algorithms after integration with the target task is complete. Apparently in these cases, allowing relearning of the source task model during integration with the target task data is indeed destructive to source task performance. LwF outperforms Fine-Tuning significantly in knowledge retention for movie reviews, but interestingly does not for the emoticon heuristic. The effect of the greedy target task initialization strategy also appears inconsistent. It seems it is possible that this greedy initialization could improve our proposed forgetting cost paradigm in some cases as well. However, a rigorous analysis of the tradeoffs for this initialization approach is beyond the scope of this paper.  

As the source task representation is literally stored fixed as part of the target task representation in progressive neural networks, it is not clear how to assess any effective forgetting of the source task during target task integration. As a result, we omit them from our source task forgetting experiments.

\begin{table}
\small
\centering
\begin{tabular}{|l|c|c|c|c||c|}
\hline \bf Source Task & \bf Fine-Tuning & \bf LwF & \bf Greedy LwF & \bf Forgetting Cost & \bf Source Only \\ \hline
Binary Movie Reviews & 80.7\% & 81.3\% & 81.5\% &  \textbf{83.3\%} & 85.5\% \\ \hline
Five Class Movie Reviews & 41.6\% & 42.8\% & 43.1\% &  \textbf{43.3\%} & 45.9\% \\ \hline
Emoticon Heuristic & 59.4\% & 59.1\% & 58.9\% & \textbf{60.3\%} & 63.4\% \\ \hline
\end{tabular}
\caption{\label{tabv} Evaluation of accuracy on the source task after integration with the target task data of SemEval 2016 Task 4 Subtask A. The accuracy after only source task training prior to integration with the target task is included for reference as a baseline.}
\end{table}

\subsection{Inspection of Learned Representations}

Now that we have established the empirical benefits of our proposed forgetting cost, we will demonstrate what it achieves qualitatively through examples. In Table \ref{tab3} we include a sample of examples that are predicted correctly by transferring the knowledge source with the forgetting cost paradigm and not with fine-tuning based integration. The effect is, perhaps, easiest to understand for the rule based and movie review based transfer scenarios. For the rule based transfer setting you can literally map insights that are not forgotten to their respective logical rule in the model, as is the case in these examples. Moreover, we can see movie domain specific terminology such as "May the force be with" is seemingly forgotten with standard fine-tuning, but not when the forgetting cost regularization is applied.

Considering that we have shown a neural network can distill and improve a representation learned by a logical rule engine, how the final representation differs from the logic of the original engine is of practical interest. We thus compare the agreement of our fine-tuned rule based GRU with the original rule model on the SemEval testing set. We find that the transferred model achieves 78.7\% agreement with the rule model when the rule model is right. This clearly indicates that our final model is not deterministic based on the rule engine, and has a probability of adding errors even when the original rule model works well. However, our model actually has 44.7\% accuracy on the examples the rule model got wrong. Our approach yields significant gains in comparison to the original rule classifiers, improving from 57.8\% to 64.4\% test set accuracy before even incorporating in auxiliary knowledge sources. 

\begin{table*}[t]
\small
\centering
\begin{tabular}{|c|p{7cm}|c|c|c|}
\hline \bf Source & \bf Tweet & \bf Label & \bf Fine-Tuning & \bf Forgetting Cost \\ \hline
Logical Rules & John Kasich should feel proud of his performance at the \#GOPDebate Thursday night. He looked more presidential than the rest of the field. & Positive & Neutral & Positive \\ 
\hline
Logical Rules & @BrunoMars I'm so tired of you dressing like you ain't got no money. You went from wearing Gucci loafers to 6th grade boy Sketchers. & Negative & Neutral & Negative \\
\hline
Logical Rules & @DavidVonderhaar loving the beta Vahn, even playing it on PC with a  PS4 controller without aim assist, can't wait for November 6 & Positive & Neutral & Positive \\
\hline
Movie Reviews & Selena Gomez presented Amy Schumer with an award and a heap of praise at the Hollywood Film Awards on November 1. & Positive & Negative & Positive \\
\hline
Movie Reviews & mailjet: It's Fri...we mean Star Wars Day. May the force be with all of your emails! https://t.co/FbDdjiJVUT & Positive & Neutral & Positive \\ 
\hline
Movie Reviews & Straight Outta Compton's success hopefully convinces New Line Cinema to give Ice Cube the right budget for the last Friday movie. & Positive & Neutral & Positive \\
\hline
Emoticons & That ball Kris Bryant just hit is the 2nd farthest ball I've ever seen hit. He is officially ridiculous. & Positive & Neutral & Positive \\ \hline
Emoticons & This fandom's a mess omg, I wouldn't be surprise if tomorrow there's a trend who says Niall's going to marry his cousin \#WeKnowTheTruth & Negative & Positive & Negative \\ \hline
Emoticons & Christians snapchat story makes me want to kill myself..like I feel like a depressed 8th grader going through that emo phase & Negative & Neutral & Negative \\ 
\hline
\end{tabular}
\caption{\label{tab3} Some transfer learning examples from each knowledge source to SemEval 2016 where the GRU model successfully predicts sentiment when using the forgetting cost paradigm, but not with fine-tuning based integration.}
\end{table*}

\section{Integrating Transfer Learning from Multiple Tasks with Ensemble Distillation}

\subsection{Ensemble Methodology}

In our experiments we tried to find a balance between an ensemble model that is powerful enough to have an adaptive weighted average decision function and not so powerful that it overfits on our limited training and validation data. Our model is quite similar in architecture to the gating network component of a hierarchical mixture of experts model \citep{HMEHinton}, \citep{HMEJordan}. We tried our model over all four representations at once and found that it overfits. Our experiments showed it is more effective to adopt a greedy ensembling strategy where all models are combined with the best performing model on the validation set at each phase until only two models are left. Finally, these two models are combined with the same mechanism. \citep{Riemer16} suggests that a many element gating network can be improved with a sparsity constraint, but this did not work as well as the greedy strategy for our model and experiments. 

More formally, for any two models $A$ and $B$ combined in an ensemble, we train the following mechanism using Stochastic Gradient Descent:

\begin{equation}
m_A = \sigma(W_{A} \hat y_A + b_{A})
\label{eq:ma}
\end{equation}
\begin{equation}
m_B = \sigma(W_{B} \hat y_B + b_{B})
\label{eq:mb}
\end{equation}
\begin{equation}
a_{A} =  \frac{m_{A}}{m_A + m_B}  \label{eq:aa}
\end{equation} 
\begin{equation}
a_{B} =  \frac{m_{B}}{m_A + m_B}  \label{eq:ab}
\end{equation} 
\begin{equation}
\hat y_{ensemble} = a_A \hat y_A + a_B \hat y_B 
\label{eq:ye}
\end{equation}
where $\hat y_{ensemble}$ is the prediction vector of the combined ensemble. $\hat y_A$ and $\hat y_B$ are the output vectors of the individual models. 

\subsection{Ensemble Results}

Our ensemble model was trained on what was set aside as the validation data during the initial training with early stopping. In the first phase of combining, the model transferred from the logical rule source task was combined with each model. In the second phase, the model based on transfer from the binary movie review sentiment model was combined with each model. In the third phase, the two remaining models were combined. The results of our ensemble in Table \ref{tab4} suggest that it is possible to further improve the performance of a single sequential transfer model by intelligently combining its predictions with models that have other perspectives. This is because they are modeled using different source tasks for prior knowledge. Impressively, our final distilled model surpasses results from all prior models on the SemEval 2016 benchmark using the same final architecture of a 50 hidden unit GRU model that is clearly not even competitive when trained simply on the task specific labeled data. The prior best model SwissCheese \citep{SwissCheese} consists of random forests ensemble built utilizing multiple convolutional neural network models and distant supervision. In fact, we achieve superior results despite using over an order of magnitude less total data for training our model.

We would also like to underscore that our total improvement of 1.5\% as a result of creating an ensemble with our best transferred model from the logical rule source task can be viewed as quite disappointing, despite achieving state of the art results. In fact, in the theoretical limit of having a decision model that switches to the best already learned model at each point, our four transferred representations would achieve 85.1\% accuracy together. For the combination of the movie review based models and logical rule based model we can get to 81.4\% accuracy. Moreover, we can get 76.5\% accuracy with just the logical rule based transfer model and the emoticon prediction based transfer model. Unfortunately, we achieve nowhere near these theoretical results despite representations that are apparently quite diverse. This seems indicative that there are significant gains yet to be uncovered in integrating these representations. 

\begin{table}
\small
\centering
\begin{tabular}{|l|c|}
\hline \bf Model Description & \bf Accuracy on SemEval Test Set  \\ \hline
Distilled GRU Trained on Full Ensemble & \bf 66.0\% \\  
Full Ensemble & 65.9\% \\ 
Ensemble with Logical Rules and Both Movie Review Tasks & 65.7\% \\ 
Ensemble with Logical Rules and Binary Movie Reviews & 65.4\% \\ 
Ensemble with Logical Rules and Five Class Movie Reviews & 65.1\% \\ 
Ensemble with Logical Rules and Emoticon Prediction & 65.0\% \\
Ensemble with Both Movie Review Tasks & 62.1\% \\
GRU Trained on Only SemEval Data & 53.6\% \\
\hline
SwissCheese \citep{SwissCheese} & 64.6\% \\
NTNUSentEval \citep{No2} & 64.3\% \\
UniPI \citep{Unipi} & 63.9\% \\
CUFE \citep{CUFE} & 63.7\% \\
INSIGHT-1 \citep{insight} & 63.5\% \\
\hline
\end{tabular}
\caption{\label{tab4} Empirical three way sentiment classification results on the SemEval 2016 Task 4 Subtask A test set.}
\end{table}

\section{Conclusion}

We consider a new methodology called the forgetting cost for preventing the catastrophic forgetting problem of neural network sequential transfer learning. The forgetting cost is practical and easy to implement. We have demonstrated for the challenging task of Twitter sentiment analysis that it can uncover significant gains in generalization performance and that it seems to not forget knowledge traditionally forgotten from the source task during fine-tuning. Our strong empirical results still motivate multiple avenues with high potential for continued exploration in text analytics. Using logical rules to improve neural network models is a promising direction for humans to efficiently contribute to increased model performance. Additionally, the large diversity of representations learned from multiple classifiers with the same target task but different source tasks seems to indicate there is potential to see even much greater gains when integrating multiple sources of knowledge transfer. 

\bibliography{iclr2017_conference}
\bibliographystyle{iclr2017_conference}

\appendix

\section{Mapping Sentiment Rules to Soft Targets}
\label{AppendixA}

The gazetteer based logical rule engine separates sentences and phrases in the text. It then applies dictionaries of positive and negative sentiment words and phrases to the corresponding text.  For each positive or negative phrase found, it checks to see if negation or double negation are applied, and modifies the polarity of the sentiment accordingly. The result for any piece of text is a count of positive and negative sentiment occurrences. For this task, we simply count the total number of positive and negative indicators to give an overall positive, negative or neutral score. To be concrete, we have a simple procedure for mapping positive and negative word counts to soft labels that could be used for distillation. If there are no positive or negative words, the output vector is a one hot vector corresponding to a neutral label. If there are an unequal number of positive and negative sentiment words, the neutral label is zero and the raw counts are sent to the softmax function to create a soft label over the positive and negative word occurrences. Finally, if there are an equal amount of positive and negative words, we consider the added total sentiment words plus one in the neutral label as well as the number of positive words and negative words before sending these totals through a softmax function. 

\section{Size Selection for the Rule Distillation Task}
\label{AppendixB}

\begin{table}
\small
\centering
\begin{tabular}{|c|c|c|c|}
\hline \bf Hidden Units & \bf Examples & \bf Alignment with Teacher & \bf Accuracy on SemEval Test Set  \\ \hline
25 & 50,000 & 88.3\% & 59.1\% \\
25 & 300,000 & 91.9\% & 58.6\% \\
\hline
50 & 50,000 & 88.6\% & 58.9\% \\
50 & 300,000 & 93.0\% & 58.5\% \\
\hline
75 & 50,000 & 88.7\% & 58.9\% \\
75 & 300,000 & 93.6\% & 58.3\% \\
\hline
100 & 50,000 & 88.6\% & 58.7\% \\
100 & 300,000 & 93.8\% & 58.1\% \\
\hline
125 & 50,000 & 88.5\% & 58.7\% \\
125 & 300,000 & 93.7\% & 58.3\% \\
\hline
150 & 50,000 & 88.5\% & 59.0\% \\
150 & 300,000 & 94.0\% & 58.5\% \\
\hline
\end{tabular}
\caption{\label{tab2} Logical rule engine distillation performance and SemEval 2016 Task 4 Subtask A accuracy as a function of the number of hidden units in the GRU and the number of training examples. The 50 hidden unit and 50,000 training example model performs the best on the SemEval training set.}
\end{table}

In Table \ref{tab2} we detail the performance of distilling a logical rule engine into a GRU based recurrent neural network by imposing soft labels over unlabeled tweets. The fact that we keep our word representations fixed with general purpose unsupervised data makes it difficult for the GRU to distill the entire model without a large number of examples. Additionally, as there were a large number of examples in our distillation experiments, we did not experience high run to run variation and only trained a single GRU model for each distillation experiment (as opposed to picking the best validation error of 10 parallel training routines as in our transfer experiments). Our distilled GRU is better on the testing set than the original classifier, likely because this input representation prevents the model from overfitting to the idiosyncrasies of the rule engine. This actually underscores an important point for the distillation of abstract knowledge. If the target task is known during distillation, it may be beneficial to stop short of totally distilling the original knowledge as it may hurt down stream performance past a certain point. We impose a simple policy where the best hidden unit and training example combination is selected based on performance on the training data of the target task. As a result, we use the model with 50 hidden units based on 50,000 training examples in our experiments integrating with other knowledge. This model is a pretty good one to choose, and achieves high transfer performance relative to models that overfit on the teacher network.

\end{document}